\documentclass[10pt,twocolumn,letterpaper]{article}

\usepackage[pagenumbers]{cvpr} 

\usepackage{graphicx}
\usepackage{amsmath}
\usepackage{amssymb}
\usepackage{bm}
\usepackage{booktabs}

\usepackage{multirow}
\usepackage{xcolor}
\usepackage{enumitem}

\usepackage{algorithm}
\usepackage{algpseudocode}
\usepackage{caption}
\usepackage{subcaption}

\usepackage[accsupp]{axessibility}

\usepackage{cuted}

\usepackage[pagebackref,breaklinks,colorlinks,bookmarks=false]{hyperref}

\usepackage[capitalize]{cleveref}
\crefname{section}{Sec.}{Secs.}
\Crefname{section}{Section}{Sections}
\Crefname{table}{Table}{Tables}
\crefname{table}{Tab.}{Tabs.}

\def\cvprPaperID{7060} 
\def\confName{CVPR}
\def\confYear{2022}

\begin{document}

\title{Explaining Deep Convolutional Neural Networks \\ via Latent Visual-Semantic Filter Attention}

\author{Yu Yang, Seungbae Kim, and Jungseock Joo\thanks{To appear in CVPR 2022 (oral presentation).}\\
University of California, Los Angeles\\
{\tt\small yuyang@cs.ucla.edu, sbkim@cs.ucla.edu, jjoo@comm.ucla.edu}
}
\maketitle

\begin{abstract}
   Interpretability is an important property for visual models as it helps researchers and users understand the internal mechanism of a complex model. However, generating semantic explanations about the learned representation is challenging without direct supervision to produce such explanations. We propose a general framework, Latent Visual Semantic Explainer (LaViSE), to teach any existing convolutional neural network to generate text descriptions about its own latent representations at the filter level. Our method constructs a mapping between the visual and semantic spaces using generic image datasets, using images and category names.  It then transfers the mapping to the target domain which does not have semantic labels. The proposed framework employs a modular structure and enables to analyze any trained network whether or not its original training data is available. We show that our method can generate novel descriptions for learned filters beyond the set of categories defined in the training dataset and perform an extensive evaluation on multiple datasets. We also demonstrate a novel application of our method for unsupervised dataset bias analysis which allows us to automatically discover hidden biases in datasets or compare different subsets without using additional labels. The dataset and code are made public to facilitate further research.\footnote{\url{https://github.com/YuYang0901/LaViSE}}
\end{abstract}

\section{Introduction}
Convolutional neural networks have shown great performance in visual representation learning, but the learned representations are usually hard to explain or interpret. The lack of explainability raises the concern that AI systems and models, although very accurate in prediction, may have hidden negative effects on human users and society, such as AI bias. Several studies reported biases in computer vision models for face attribute classification~\cite{buolamwini2018gender,das2018mitigating}, recognition~\cite{kortylewski2018empirically,wang2020mitigating}, and image captioning~\cite{hendricks2018women}. It is very challenging, however, to identify these biases from a black-box model with distributed knowledge. 

\begin{figure}[t]
\centering
\vspace{-15pt}
\includegraphics[width=0.9\columnwidth]{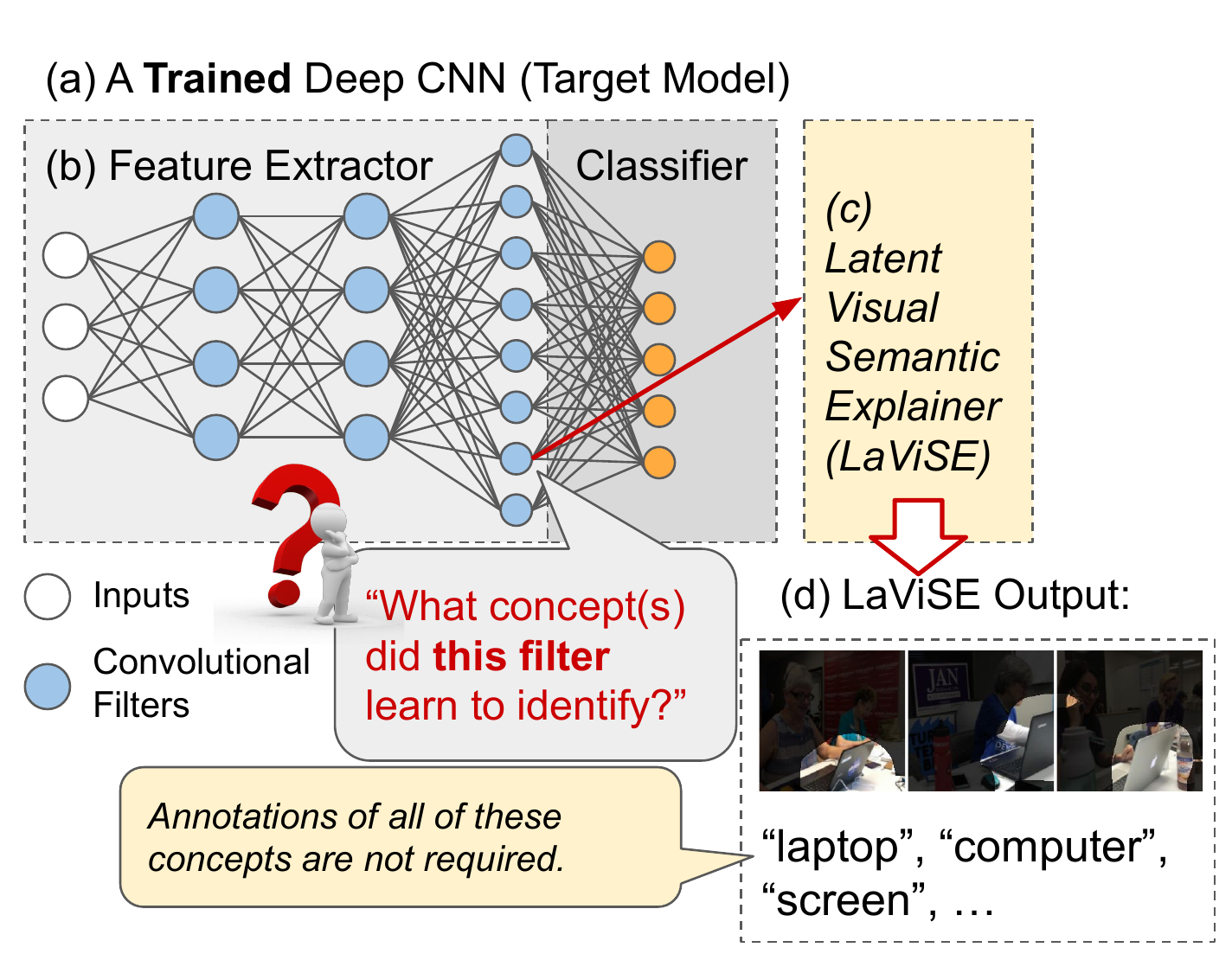} 
\vspace{-10pt}
\caption{The proposed framework aims to semantically explain the concepts learned by individual filters in a CNN without  supervision on the concepts used for the semantic explanations.}
\label{fig:intro}
\vspace{-15pt}
\end{figure}

To date, several methods have been proposed to interpret what are learned and captured in CNNs. These methods vary greatly by the form (visualization, captions, synthesized samples), the focus (individual filter vs network level), and the scope (any existing models vs requiring training with supervision) of the generated explanations, and each method has its own strengths and weaknesses. 

The main objective of this paper is to generate the textual interpretations of any existing black box model that can overcome the limitations of existing approaches for several reasons. First, it generates words and thus can be more semantically meaningful and objective than visualization based methods~\cite{zeiler2014visualizing,selvaraju2017grad,olah2017feature}. 
Second, it can apply to any arbitrary network and does not require training or annotations, which is much more applicable than methods that require training a model with ground-truth explanation annotations~\cite{huk2018multimodal}. We do train an adapter using general image classification datasets which can apply to any given target model. 
Third, it can generate explanations using novel concepts that are not given in the training set. 
These properties are critical in understanding black-box models for which we do not have access to the original training data or any information about the training process. This is a realistic assumption in practice where one needs to interpret and scrutinize a given model.

To this end, we introduce the \emph{Latent Visual Semantic Explainer} (LaViSE) as a novel framework to teach any existing CNN to generate text descriptions about its own latent representations at the filter level. Our framework differs from supervised approaches in that we do not require to annotate the explanation itself along with an input image and a category label. Instead, our method constructs a mapping between the visual and the semantic space using generic image datasets (using images and category names), then transfers the mapping to the target domain without semantic labels. 
We do not attempt to train more ``interpretable'' models~\cite{zhang2018interpretable, nauta2021neural, bohle2021convolutional, liang2020training, d2021ganmut} but interpret any given network without changing its structure or retraining. Our work is also closely related to the literature of visual attribute or concept based learning~\cite{berg2010automatic,rastegari2012attribute,patterson2014sun,ge2021peek,shen2020interpreting}, but we do not require any additional supervision for attribute labeling, which makes our approach more generally applicable. 
It is also important to note that our method does not merely explain each individual filter separately but uses aggregated responses using a novel filter attention method. 
Experimental results show our method can generate novel descriptions for learned filters beyond the set of categories defined in the training dataset and provide more accurate explanations for filters comparing to the existing method.

While our main contribution is a novel method to generate explanations for any CNNs, our approach can be used in a practical application of comparative analysis where the goal is to discover and explain the differences between given multiple models or multiple sets of images. To demonstrate the utility, we compare a model finetuned from a pretrained model and a model trained from scratch, and we also compare the gender disparities in datasets. Besides public image datasets, we collect and analyze social media photographs posted by U.S. politicians to exemplify the effectiveness of our method in solving more challenging real-world problems. 


\section{Related Work}\label{sec:related}
\noindent
\textbf{Explanations via Visualization.}
Saliency methods \cite{zeiler2014visualizing,simonyan2013deep,selvaraju2017grad,oramas2018visual,gu2021interpreting,Huang_2020_CVPR} visually show the amount of contribution for each pixel to the model prediction. They are widely used in the literature but may be unreliable \cite{ghorbani2019interpretation} because they can respond to low level features such as image edges rather than semantically more important features~\cite{adebayo2018sanity}. They also require users to speculate meanings of the visualizations as they do not provide explicit semantic explanations. Some approaches have been proposed to perform case-based reasoning \cite{chen2019looks,li2018deep} by providing patches from training images as explanations, i.e. by analogy. These methods cannot be applied to arbitrary networks across domains. 

\noindent
\textbf{Explanations by Text.}
\cite{hendricks2016generating} propose to generate text to explain and justify the output of an image classifier. Similarly, \cite{huk2018multimodal} take a hybrid approach and generate multimodal explanations by using both visual highlights and textual descriptions. \cite{wu2018faithful} propose a VQA system that can not only provide multimodal explanations but also link terms in the textual explanation and segmented items in the image. 
\cite{ye2019interpreting} use multimodal cues to interpret hidden messages (why and what) in visual advertisements. 
These methods can generate a very meaningful and interpretable explanation to human users, however the explanation itself should be labeled for each example, and the model will learn to generate it in the same way it computes its outputs. Also, since these explanations are annotated by humans before training, they do not necessarily explain what the model has learned (different networks will yield the same explanations). 

\noindent   
\textbf{Visual Semantic Explanations of Visual Representation.}
Our paper is the most closely related to~\cite{zhou2014object,bau2017network,rombach2020making}. They explain an individual internal filter of a trained neural network by measuring the alignments of images' activations on that filter to each predefined concept's segmentation masks. Furthermore, \cite{bau2017network} provides a dataset with a broad range of concepts annotated for the alignments. 
Our framework also uses an annotated dataset so the model can learn to construct the mapping between the visual representation and semantic embedding space, but it differs in that it can discover unseen novel concepts in another domain instead of being restricted to the annotated ones in training data.

\noindent
\textbf{Generalization to Unseen Subjects. }
Zero-shot learning (ZSL) aims to recognize instances of categories that have not been seen in the training stage. Handcrafted attributes and semantic representations learned from textual data are often used to connect the seen and unseen classes, so knowledge learned from training classes can be transferred to unseen categories. Most existing deep neural network based ZSL works 
\cite{frome2013devise,socher2013zero,norouzi2013zero,lei2015predicting,zhang2015zero,kodirov2015unsupervised,kodirov2017semantic,annadani2018preserving} either use the semantic space or an intermediate space as the embedding space. Our work relates to this topic as it also helps to discover unannotated concepts. We do not compare our framework with results in this area because our framework is not designed for ZSL but can be built on any ZSL methods.

\begin{figure*}[t!]
     \centering
     \begin{subfigure}[b]{0.45\textwidth}
         \centering
         \includegraphics[width=\textwidth]{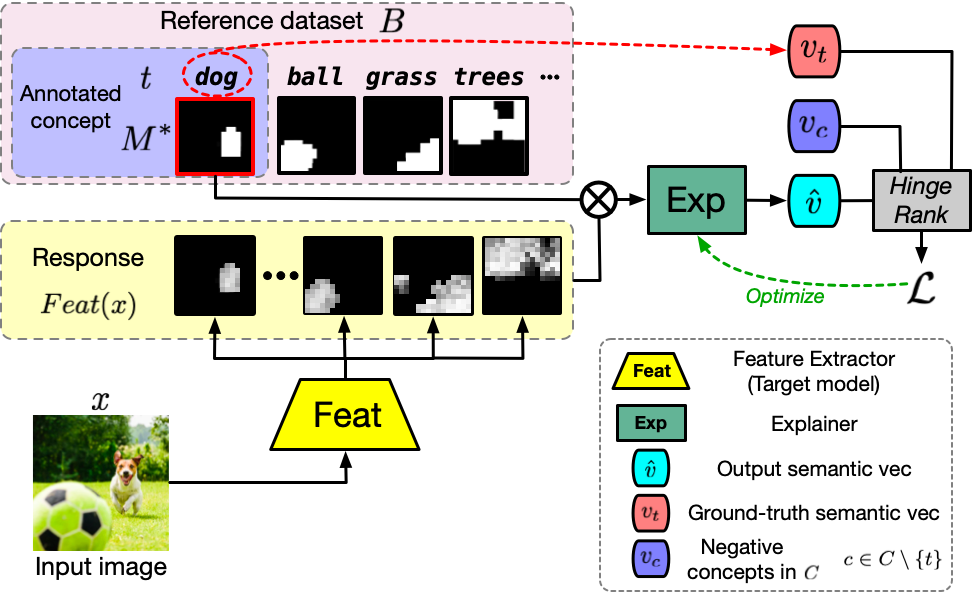}
         \caption{Training the explainer (Section~\ref{sec:exp})}
         \label{fig:LaViSE_training}
     \end{subfigure}
     \hfill
     \begin{subfigure}[b]{0.45\textwidth}
         \centering
         \includegraphics[width=\textwidth]{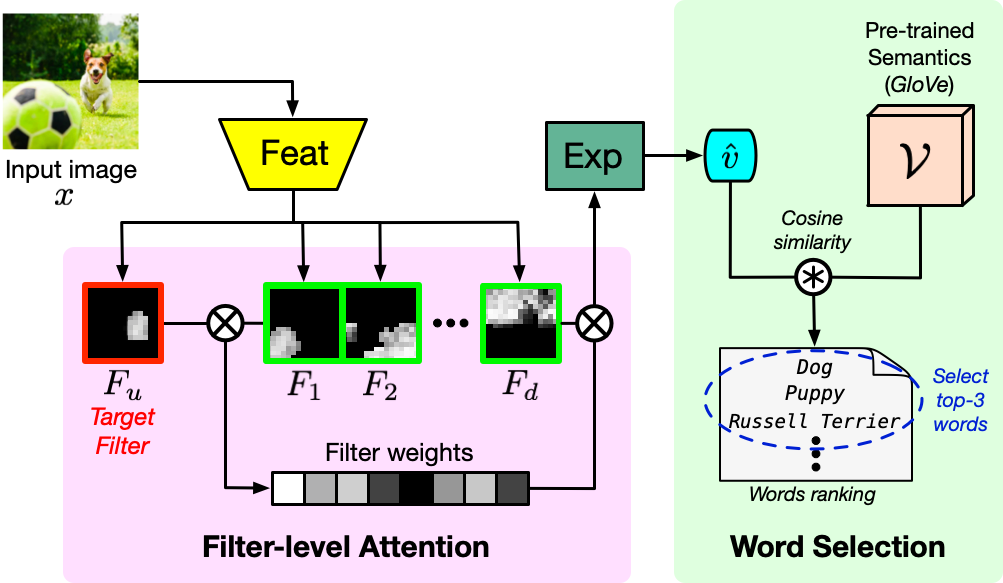}
         \caption{Inference with the filter attention (Section~\ref{sec:inf})}
         \label{fig:LaViSE_inference}
     \end{subfigure}
        \caption{An overview of LaViSE framework. (a) At the training phase, we train the explainer by connecting each image's visual representation with semantic concepts. The hinge rank loss helps the explainer learn the semantic embedding vectors, $\hat{v}$, that are close to the ground truth concept vectors $v_t$, while being far away from all the other concept vectors $v_c$ in the semantic space. (b) During inference, LaViSE obtains the representation for each latent filter of the target layer via filter-level attention, and then the trained explainer takes this representation to explain these filters semantically by selecting words with the highest similarities from $\bm{\mathcal{V}}$.}
        \label{fig:LaViSE}
\vspace{-5pt}
\end{figure*}


\section{Latent Visual Semantic Explainer (LaViSE)}
\label{sec:method}
We now explain our main framework to explain the deep visual representations of a given \textit{target model} (a CNN), $f$. We assume that this model has already been fully trained from an unknown target dataset, $D$, and we do not have access to $D$. In order to learn the visual-semantic explainer on $f$, we instead use another dataset, the \textit{reference dataset} $B = \{(x_i, y_i)\}^{n}_{i=1},$ where $x_i \in \mathbb{R}^{3 \times h\times w}$ is an input image and $y_i$ is a set of concept labels associated with the image and corresponding masks, \ie $y_i = \{(t_j, M_j)\}^{m}_{j=1}$. $t_j \in C$ is an annotated concept, and $C$ is the set of all concept labels in $B$. 
$M_j \in \{0, 1\}^{h\times w}$ is a pixel-level mask for its corresponding concept, $t_j$. The reference dataset may already provide these masks (e.g. semantic segmentation). For object detection datasets, the region inside each bounding box will be filled in with 1. For image classification datasets (no bounding boxes),  $M_j=1^{h\times w}$. For each semantic concept category $t_j$, we obtain its semantic representation, $v_{t_j}$ by using a pre-trained word embeddings (e.g. GloVe~\cite{pennington2014glove}).

In order to explain filters in an arbitrary target model, LaViSE first trains a feature explainer (Section \ref{sec:exp}) using a reference dataset, $B$. Once trained, this explainer is further used to explain filters that may not have any matching concepts in $B$ using our novel filter attention mechanism (Section \ref{sec:inf}).

\subsection{Training  Feature Explainer}
\label{sec:exp}
The purpose of our feature explainer is to transform the visual feature representations of images from a target model to equivalent representations in a semantic space, which can be translated to words. To this end, we train a  feature explainer $\text{Exp}(\cdot)$ with a reference dataset $B$ as shown in Figure~\ref{fig:LaViSE_training}. Given a target model, $f$, we take the feature extractor $\text{Feat}(\cdot)$ (\eg by removing the last classification layer). We use the feature extractor to obtain the visual representations of the images in $B$, $F_i = \text{Feat}(x_i) \in \mathbb{R}^{d \times h'\times w'}$, \ie the output of the last convolutional layer with $d$ filters. This is the input to the feature explainer, but we first mask this feature as follows. Each image has $k$ pairs of a ground-truth concept and its mask, $(t_j, M_j)$\footnote{$M_j$ is resized to the size of feature response map ($h' \times w'$).}. For each concept, we obtain a masked visual feature by element-wise product, $F_i \odot M_j$, because this masked feature corresponds to the specific concept, $t_j$. We then use the feature explainer to obtain the semantic representation of the masked visual representation: $\hat{v}=\text{Exp}(F_i \odot M_j)$. 

To train the explainer, we use the semantic representations of the ground-truth concept, $v_{t_j}$, and negative concepts, \{$v_c:  \forall c \in C, c \neq t_j$\}, using a pre-trained word embedding model such that $\hat{v}$ should be closer to $v_{t_j}$ than to $v_c$. Note that we always normalize these semantic vectors. 
We modify the objective function from~\cite{frome2013devise} as follows:
{\small
\begin{align}
    \min_\theta \frac{1}{k^*} \sum_j \sum_{c\neq t_j} \text{max}(0,1-v_{t_j}^\intercal\hat{v}+v_{c}^\intercal\hat{v}). \label{eq:objective}
\end{align}}%
Our objective function combines dot-product similarity and hinge rank loss as this combination has shown better performance than other losses~\cite{xian2017zero} in zero-shot learning. 
As LaViSE is a general framework developed for the filter-level interpretability, it can incorporate any other user preferred loss functions or additional loss terms that can serve the purpose of training a zero-shot mapping from visual representations to semantic embeddings.
The procedure of training the explainer is shown in Algorithm~\ref{alg:algorithm_lavise}.

\begin{algorithm}[t]
\caption{LaViSE Training}\label{alg:algorithm_lavise}
{\small
\begin{algorithmic}
\State {\bfseries Input:} Target feature extractor $\text{Feat}(\cdot)$,
reference dataset $B$ and the set of all annotated concepts $C$, pretrained semantic embedding $\mathcal{V}$. 
\State {\bfseries Output:} Trained explainer $\text{Exp}(\cdot)$ with parameters $\theta_{exp}$
\State Initialize explainer $\theta_{exp}$
\For{each image $x$ and its annotations $\{(M, t)\}$ in $B$}
\State Get features $F \leftarrow \text{Feat}(x)$
\For{each mask-concept pair $(M_j, t_j)$ of $x$}
\State Get explainer output $\hat{v} \leftarrow \text{Exp}(F\odot M_j)$
\State Get ground-truth concept embedding $v_{t_j} \in \mathcal{V}$
\For{each concept $c\in C\setminus\{t_j\}$}
\State Get concept embedding $v_c \in \mathcal{V}$ 
\State $l \leftarrow l + \text{max}(0,1-v_{t_j}^\intercal\hat{v}+v_{c}^\intercal\hat{v})$ 
\EndFor
\EndFor
\State Update $\theta_{exp}$ by Adam to minimize $l$ (eq~\ref{eq:objective}). 
\EndFor
\end{algorithmic}} 
\end{algorithm} 

\subsection{Inference with  Filter Attention}
\label{sec:inf}
The goal of training and using the feature explainer is to explain filters that describe concepts not specified in the reference dataset. For example, the dataset may contain ``football'' as a concept but lack other related concepts such as ``stadium'' or ``referee'' which are still likely present in the images. Since the feature explainer learns to map any visual features to a semantic space, it can generalize to discover \textbf{novel} concepts. 

A naive way to use the explainer for each filter is to only use the activation of the filter and suppress other filters' responses. We found that this naive approach leads to very poor performance, and this may be due to the fact that many filters collectively capture visual features in a distributed manner. This suggest that we can still use the entire feature responses, with proper modification, even when explaining one filter. Some previous studies in interpreting filters have tried to use masking on filter activations~\cite{zhang2018interpretable} and use all the filters together. We propose a novel attention-based masking mechanism, which is simple but effective in extracting and reweighting features relevant to each target filter. 
Our method is similar to recent self attention models, \eg Transformer~\cite{dosovitskiy2020image,vaswani2017attention}, but simpler because it doesn't use repetitive blocks or multi-head attention.  

\begin{figure}[t]
  \centering
  \includegraphics[width=\linewidth]{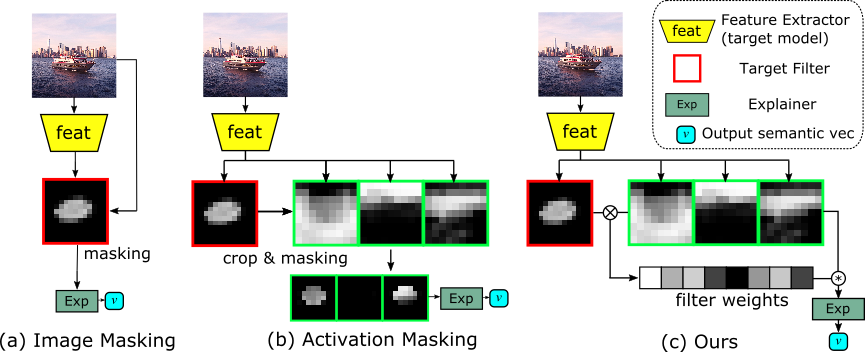}
   \caption{An illustration of comparing our filter attention method with other masking methods.}
\vspace{-15pt}
   \label{fig:baseline_methods}
\end{figure}

\noindent\textbf{Filter-level Attention.}
To discover important concepts which are implicitly represented and distributed over many filters, we propose the filter attention module. Instead of using each filter's activation separately, our method finds a representation for each filter using activations of all filters collaboratively via an attention-based approach.
Essentially, we take advantage of the spatial alignments between filters describing a concept and collect their focused activations.
In our filter attention method, the feature response map of a target filter ($u$, the filter we want to explain) serves as a spatial attention; the other filters are reweighted based on their similarities to the target filter. 
Specifically, suppose we have an input image, $x$, and $d$ filters in the feature extractor, $\text{Feat}(\cdot)$, and the computed feature response map is 
$F = [F_1, F_2, ..., F_d] = \text{Feat}(x) \in \mathbb{R}^{d \times h' \times w'}$. 
The input to the explainer, $Exp(\cdot)$, with respect to a target filter $u$, is computed as follows: $F^{\text{att}}_k = a ( F_u, F_k ) \cdot F_k, \forall k$, where $a(\cdot)$ computes spatial correlations between filters by cosine similarity.
Figure~\ref{fig:baseline_methods} illustrates the difference between our method and other baseline methods.


\noindent\textbf{Word Selection Method.}
To obtain a list of words to explain given images, LaViSE next computes the cosine similarities between each semantic filter embedding vector and all words in $\mathcal{V}$, and collect $s$ words with the highest similarities. For each filter, we gather $s\times p$ words collected from the top $p$ most activated images and rank the words based on their frequencies. Finally, our framework composes explanations for filters using top-x ranked words. The users can decide the number of words used for each filter explanation depending on how much detail they desire. Note that we empirically tested choices of parameters $s$ and $p$, and found that our framework works well when $p$ is an integer between $5$ and $25$ while the choice of $s$ may depend on the size of $\mathcal{V}$ and does not have a substantial impact on the results.

\section{Experiments}
\subsection{Datasets}
To evaluate our proposed framework, we use the following three publicly available datasets and one novel dataset that we collected, which are used as a target dataset, a reference dataset, or both:
\textbf{(i) MS COCO \cite{lin2014microsoft}.} MS COCO 
has more than 200K daily scene images that are pixel-wise labelled with 80 common object categories. To help with our analysis, we also include the gender annotations of MS COCO provided by \cite{zhao-EtAl:2017:EMNLP20173}.
\textbf{(ii) Visual Genome (VG) \cite{krishna2017visual}.} We only use images that have box-able annotations for our experiments. During pre-processing, we combined object categories based on their synset names, combined instances of the same category in the same image, and deleted categories that appear in less than 100 training images. In the end, there remains 106,215 out of 107,228 images,  1,208 out of 80,138 object categories, and at most 47 object categories per image.
\textbf{(iii) Broden \cite{bau2017network}.} Broden combines selected examples from several densely labeled image datasets to provide pixel level ground truth labels to a broad range of visual concepts, including scene, objects, object parts, textures, and materials. NetDissect~\cite{bau2017network} leverages this dataset to provide explanations so we use it to directly compare our framework with NetDissect without biases imposed by the choice of datasets.
\textbf{(iv) Social Media Photographs of US Politicians (PoP).}
To demonstrate the practice of our framework in real-world applications, we also composed a new dataset of social media posts from accounts of US politicians to be used as a target dataset without any visual category labels. We collected roughly 80k images for politicians who ran for the 2018 election from Facebook and annotated the images by the gender and political party.
Note that, we used \textbf{GloVe \cite{pennington2014glove}} (i.e., 300d GloVe embeddings) that contains total 400K vocabularies trained on 6B Wikipedia tokens for the pre-trained word embeddings $\bm{\mathcal{V}}$.

\subsection{Setup}
For the experiments, we use ResNet~\cite{he2016deep}
as our backbone models and build our framework on the PyTorch \cite{paszke2019pytorch} implementations of ResNet-18 and ResNet-50. In Table \ref{tab:netdissect} and \ref{tab:baselines}, ``\textit{layer4}'' and ``\textit{layer3}'' refer to the module names for the PyTorch models.  

We consider two challenging settings that are the closest to the real-world scenarios: (1) the list of concepts that can appear in the target dataset is known (but still no annotation is given); (2) we have no prior knowledge about concepts that can appear in the target dataset. We imitate (1) in a generalized zero-shot learning setting with the VG dataset. We only train with a proportion of annotated classes and consider all annotated classes as all concepts that can appear in the dataset. In our experiments, we randomly selected 70\% categories for training the mapping and left 30\% categories for the model to discover. The split is manually set to ensure that we left out enough new concepts for the model to find, and meanwhile, the model can have enough supervision. We test scenario (2) with the MS COCO dataset, as it does not have as many annotated concepts as VG.


\subsection{Compared Methods} \label{sec:baselines}
To evaluate the performance of interpreting deep representation, we deploy \textbf{NetDissect}~\cite{bau2017network} as our competing method since it is the only applicable method of the filter-level approach as we discussed in Section~\ref{sec:related}. Note that explaining methods are not easily comparable as they tend to focus on unique settings.

Moreover, we carefully designed the following three baselines to show that our novel filter attention is indispensable and irreplaceable for separating representations for filters at the inference stage:\\
    \noindent\textbf{(i) Original image}: Without any attention or masking, the image goes through the feature extractor and then directly into the feature explainer. This baseline is equivalent to using a zero-shot learning model trained for image classification and then collecting the top predictions of most activated images of each filter as the explanation. \\
    \noindent \textbf{(ii) Image masking}: For each filter $u$ of layer $l$, we collect an activation map $A_l(x_i)_u$ for each $x_i\in D$, compute the distribution of all unit activations $\{a_l(x_i)_u\}_{i=1}^{n}$ on $u$ and select a threshold $T_u$ such that the probability of an activation being above the threshold is $p$, namely $P(a_u > T_u) = p$. For top activated images, we scale their activation maps of layer $l$ to the shape of the images and set regions with activations below the threshold to zero after preprocessing. We then input the masked images to the feature extractor and then the feature explainer to get the results. \\
    \noindent \textbf{(iii) Activation masking}: We use the same thresholds $T_u$'s for the image masking baseline, but we apply the masks to the activation maps directly. This method was also used in~\cite{zhang2018interpretable} for interpreting CNNs.

To ensure a fair comparison across all, we adopt the probability threshold $p = 0.005$ used in \cite{bau2017network} for all thresholds mentioned above and also for all visualizations of activated regions in this paper. 

\subsection{Evaluation Protocols}
For evaluation, we utilize two metrics~\cite{bau2017network}, perceptual effectiveness (i.e., human evaluation) and objective accuracy.

\textbf{Human Evaluation.} Since the notion of interpretability can be subjective and the standard way of quantifying interpretability is still under exploration, we first evaluate the quality of explanations with human examiners from Amazon Mechanical Turk (MTurk). For each filter, human examiners were shown 15 images from the reference dataset
with highlighted patches showing the most highly-activating regions for the filter (e.g. Fig~\ref{fig:broden}). Note that this amounts to 7-15K images used for a whole model per each setting. We have at least 3 examiners to evaluate each filter, and we take the averages of the median scores for all filters as the results, shown in Table \ref{tab:netdissect} and \ref{tab:baselines}. 
We will elaborate the protocol further in Section~\ref{result}. 

\textbf{Objective Evaluation.} We also compute the intersection-over-union (IoU) score of each annotation mask and the activation mask of each filter (i.e. segmentation). If the score is above a certain threshold, then we consider the concept corresponding to the annotation mask is one of the ground-truth concepts for that filter. We set the threshold to $0.04$ to be consistent with \cite{bau2017network}. 


\subsection{Results}\label{result}
\subsubsection{Comparing with NetDissect}

\begin{table}[tbp]
\vspace{-5pt}
\center
\footnotesize
\resizebox{\linewidth}{!}{
\begin{tabular}{ccc|ccc}
\multicolumn{3}{c|}{Settings} & \multicolumn{3}{c}{Results} \\ \hline
 Model & Target & Reference & \multirow{2}{*}{Method} & Precision & \multirow{2}{*}{Prefer} \\
 \& Layer & dataset & dataset & & (Top-1) & \\
\hline
\hline
ResNet-18 & \multirow{2}{*}{Places365} & \multirow{2}{*}{Broden} & NetDissect & 0.70 &  26\%\\
Layer 4 & & & \textbf{LaViSE} & \textbf{0.74}& \textbf{42\%} \\
\hline
ResNet-18 & \multirow{2}{*}{MS COCO} & \multirow{2}{*}{MS COCO} & NetDissect & 0.42 &  12\%\\
Layer 4 & & & \textbf{LaViSE} & \textbf{0.70}& \textbf{49\%} \\
\hline
ResNet-50 & \multirow{2}{*}{MS COCO} & \multirow{2}{*}{MS COCO} & NetDissect & 0.38 &  12\%\\
Layer 4 & & & \textbf{LaViSE} & \textbf{0.68}& \textbf{48\%} \\
\hline
ResNet-50 & \multirow{2}{*}{ImageNet} & \multirow{2}{*}{VG} & NetDissect & 0.24 &  18\%\\
Layer 4 & & & \textbf{LaViSE} & \textbf{0.46}& \textbf{42\%} \\
\hline
ResNet-50 & \multirow{2}{*}{ImageNet} & \multirow{2}{*}{VG} & NetDissect & 0.20 &  12\%\\
Layer 3 & & & \textbf{LaViSE} & \textbf{0.26}& \textbf{28\%} \\
\end{tabular}}
\caption{Human evaluation of comparing explanations generated by LaViSE (ours) and NetDissect~\cite{bau2017network} in different settings.}
\label{tab:netdissect}
\vspace{-15pt}
\end{table}

\begin{figure*}[t]
\centering
\vspace{-10pt}
\includegraphics[width=0.85\textwidth]{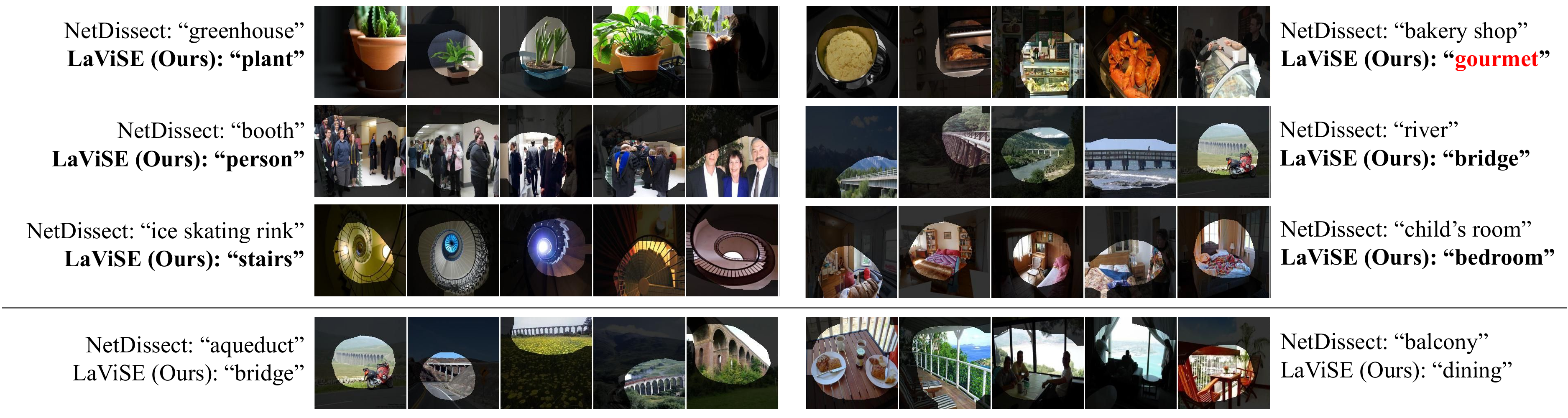}
\vspace{-8pt}
\caption{Qualitative comparison between explanations of the same filters given by LaViSE and NetDissect. We used the model (a ResNet-18 trained on Places365) and dataset (Broden) provided by \cite{bau2017network}. The first three rows are examples where human raters prefer our explanations over NetDissect's. The last row are examples where NetDissect's explanations were more favored by human raters than ours.}
\vspace{-5pt}
\label{fig:broden}
\end{figure*}

\begin{table*}[!t]
\center
\footnotesize
\scalebox{0.84}{
\begin{tabular}{ccc|ccccccc}
\multicolumn{3}{c|}{Settings} & \multicolumn{7}{c}{Results} \\ \hline
Model & Target & Reference & \multirow{2}{*}{Method} & Precision-\textit{H} & Precision-\textit{H} & Recall-\textit{H} & Recall & Recall & Recall\\
\& Layer & dataset & dataset & & (Top-1) & (Top-5) & (Top-5) & (Top-5) & (Top-10) & (Top-20)\\
\hline
\hline
\multirow{4}{*}{\shortstack[c]{ResNet-18\\Layer 4}} & \multirow{4}{*}{MS COCO} & \multirow{4}{*}{MS COCO} & Original image & 0.66 & 0.300 & 0.626 & 0.599 & 0.641 & 0.675\\
& &  & Image masking  & 0.61 & 0.280 & 0.586 & 0.567 & 0.619 & 0.659\\
& &  & Activation masking~\cite{zhang2018interpretable} & 0.68 & 0.310 & 0.658 & 0.629 & 0.676 & 0.721\\
& &  & \textbf{Filter attention (Ours)} & \textbf{0.70} & \textbf{0.320} & \textbf{0.670} & \textbf{0.675} & \textbf{0.728}& \textbf{0.776}\\
\hline
\multirow{4}{*}{\shortstack[c]{ResNet-18\\Layer 4}} & \multirow{4}{*}{ImageNet} & \multirow{4}{*}{\shortstack[c]{Visual\\Genome}} & Original image & 0.02 & 0.056 & 0.024 & 0.182 & 0.251 & 0.334\\
& &  & Image masking & 0.00 & 0.016 & 0.006 & 0.181 & 0.253 & 0.337\\
& &  & Activation masking~\cite{zhang2018interpretable} & 0.34 & 0.316 & 0.149 & 0.235 & 0.309 & 0.382\\
& &  & \textbf{Filter attention (Ours)} & \textbf{0.44} & \textbf{0.340} & \textbf{0.159} & \textbf{0.273} & \textbf{0.353}& \textbf{0.429}\\
\hline
\multirow{4}{*}{\shortstack[c]{ResNet-50\\Layer 4}} & \multirow{4}{*}{ImageNet} & \multirow{4}{*}{\shortstack[c]{Visual\\Genome}} & Original image & 0.10 & 0.096 & 0.139 & 0.160 & 0.229 & 0.302\\
& &  & Image masking  & 0.05 & 0.036 & 0.049 & 0.084 & 0.134 & 0.213\\
& &  & Activation masking~\cite{zhang2018interpretable} & 0.37 & 0.264 & 0.557 & 0.199 & 0.264 & 0.333\\
& &  & \textbf{Filter attention (Ours)} & \textbf{0.46} & \textbf{0.274}  & \textbf{0.583} & \textbf{0.226} & \textbf{0.302} & \textbf{0.373}\\
\hline

\multirow{4}{*}{\shortstack[c]{ResNet-50\\Layer 3}} & \multirow{4}{*}{ImageNet} & \multirow{4}{*}{\shortstack[c]{Visual\\Genome}} & Original image & 0.00 & 0.012 & 0.050 & 0.070 & 0.097 & 0.136 \\
& &  & Image masking  & 0.00 & 0.020 & 0.080 & 0.022 & 0.045 & 0.055 \\
& &  & Activation masking~\cite{zhang2018interpretable} & 0.24 & 0.148 & 0.470 & 0.099 & 0.155 & \textbf{0.210} \\
& &  & \textbf{Filter attention (Ours)} & \textbf{0.26} & \textbf{0.156} & \textbf{0.473} & \textbf{0.110} & \textbf{0.156} & 0.207\\
\end{tabular}}
\vspace{-5pt}
\caption{Quantitative evaluation for different masking methods. Columns with ``\textit{H}'' are the results of human evaluation. Human raters are generally giving higher scores because they also accept synonyms. Please see Section \ref{sec:masking_analysis} for the details.}
\label{tab:baselines}
\end{table*}

Table \ref{tab:netdissect} shows the quantitative comparison of our framework and NetDissect. We have various settings with different models, layers, and target and reference datasets, including the same setting from \cite{bau2017network} which uses Places365~\cite{zhou2017places} as the target dataset and Broaden as the reference dataset for a fair comparison.
The results demonstrate that LaViSE significantly outperforms NetDissect in human-evaluated top-1 precision with margins in all settings. Note that, LaViSE shows almost twice larger precision values than NetDissect in certain settings.
In addition to the precision, we ask human examiners to compare semantic explanations of LaViSE and NetDissect side by side along with the most activated images in the reference data and give a comparative rating inspired by \cite{zhou2018interpretable}. 
For each comparison, the order of the presented methods to an annotator is randomized. The ``Prefer'' column of Table \ref{tab:netdissect} records the proportions of human evaluations that prefer one method over another. The results show that human raters preferred LaViSE's explanations more often than NetDissect's with a large margin.
Figure \ref{fig:broden} provides examples of user preferred explanations.
As shown in the example images and explanations, our method provides more informative explanations than NetDissect.
It is critical to note that the key advantage of our LaViSE framework is that it can apply to any trained network to generate novel textual explanations about filters. NetDissect requires a known training dataset and hence its interpretations are limited to the annotated categories in the dataset. NetDissect, therefore, focuses on estimating the interpretability of a network architecture given a specific training dataset, whereas LaViSE can \textit{additionally} interpret any network trained from any \textbf{arbitrary unknown} data and generate \textbf{novel descriptions} beyond pre-defined categories by using a visual-semantic mapping.

\subsubsection{Comparing with Masking Schemes}\label{sec:masking_analysis}

\begin{figure}[t]
  \centering
  \includegraphics[width=0.9\columnwidth]{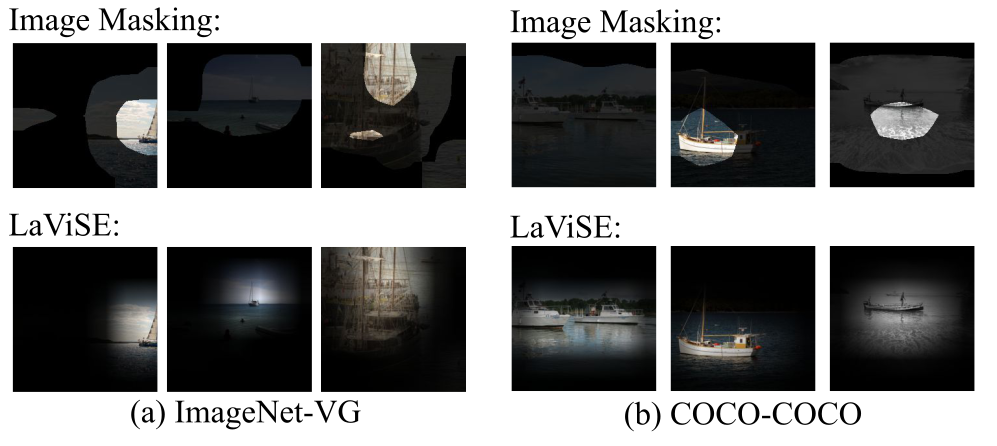}
   \vspace{-10pt}
  \caption{Qualitative examples for image masking vs. LaViSE}
   \label{fig:rep_imgs}
   \vspace{-5pt}
\end{figure}

To measure the effect of LaViSE's filter attention module, we conduct an ablation studies with three masking baseline methods described in Section~\ref{sec:baselines}.
Table~\ref{tab:baselines} shows the quantitative results when we ask human examiners to evaluate the explanations. 
Note that we conduct experiments in two different cases where one that has the same target and reference datasets, and the other with different target and reference datasets.
Since the COCO case uses the same target and reference datasets for validation purposes, all methods in the COCO case perform better than the ImageNet case where transferred features learned with ImageNet to a different reference dataset (VG).
In practice, target and reference datasets will differ since we want to explain a model already traiend with an unknown target dataset. Figure~\ref{fig:rep_imgs} showcases the example filters generated by our method and image masking method on the same images in the two cases with different datasets.

We find that our method outperforms the original image baseline and the image masking baseline with large margins in all settings, especially when we aim to explain lower layers.
This suggests that our proposed filter attention module can effectively discover important masks of given images.
Our method also outperforms the activation masking method~\cite{zhang2018interpretable} except for the top-20 recall value in the setting of ResNet-50 and layer 3.
We observe that the performance differences between these two methods are smaller in the layer 3 than the layer 4.
That is because our method is more effective in leveraging cross-filter activations in semantic mapping to capture more context at the upper layers.

\subsubsection{Explanations at Different Layers}
Different layers in a CNN may capture different types of visual concepts and also have different levels of interpretability. For example, some low level features such as texture may not be easily explainable compared to high level visual structures or objects~\cite{bau2017network}.
We observe significant performance drops at lower layers in Table~\ref{tab:baselines}.
Our human evaluation results also confirm that the lower layers of a CNN can be harder for both our framework and NetDissect to explain (Table~\ref{tab:netdissect}). To analyze the performance of LaViSE at different layers, we use a ResNet-50 pre-trained with the ImageNet as our backbone model, and Visual Genome as the reference dataset to train the explainer for each layer separately. Results in Table~\ref{tab:layer} are consistent with the human ratings.
The Supplementary Material provides more detailed analysis results due to the space limit.

\begin{table}[tbp]
\center
\footnotesize
\begin{tabular}{c|ccc}
\multirow{2}{*}{Layer} & Recall & Recall & Recall\\
  & (Top-5) & (Top-10) & (Top-20)\\
\hline
 Layer 4 & \textbf{0.226} & \textbf{0.302} & \textbf{0.373}\\
 Layer 3 & 0.110 & 0.156 & 0.207\\
 Layer 2 & 0.086 & 0.131 & 0.181 \\
 Layer 1 & 0.042 & 0.060 & 0.092\\
\end{tabular}
\vspace{-5pt}
\caption{Comparison for different conv layers of ResNet-50.}
\label{tab:layer}
\end{table}

\begin{figure}[t]
\centering
\includegraphics[width=0.75\columnwidth]{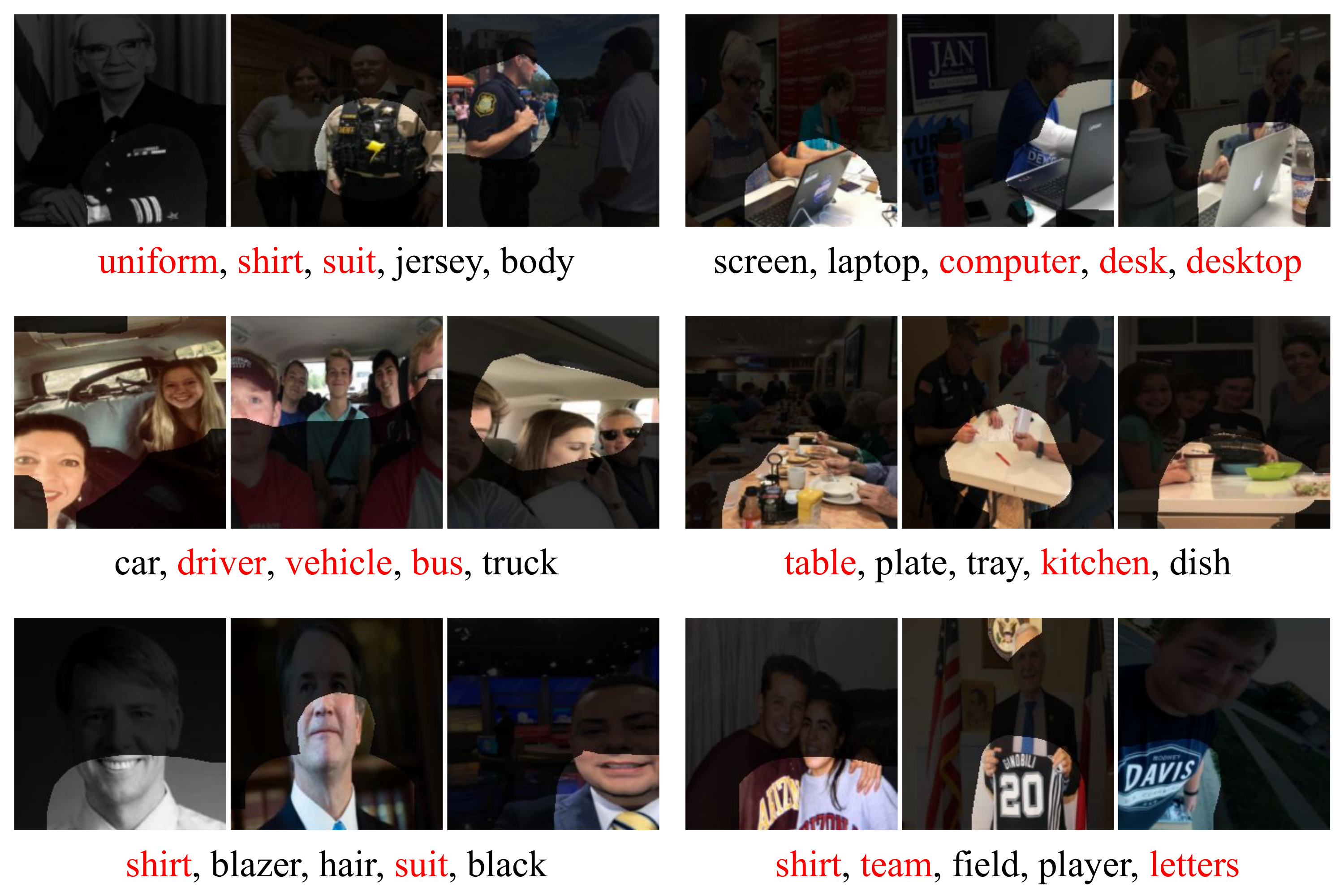}
\vspace{-10pt}
\caption{Examples of discovered concepts by LaViSE using the PoP dataset. Concepts in red are the ones outside of the pre-defined categories (i.e. no annotations) during training.} 
\label{fig:fb_new}
\end{figure}



\subsubsection{Explaining with Unsupervised Concepts}
\label{sec:new}
Our method can explain novel concepts because the semantic embedding can generalize beyond the category names given in the reference set. A similar idea has also been used in image captioning for novel objects~\cite{agrawal2019nocaps}.
We use PoP as the target dataset and Visual Genome as the reference dataset, and select 70\% categories from the Visual Genome for training the mapping, and leave 30\% categories for the model to discover. Figure~\ref{fig:fb_new} shows the examples of concepts that were discovered by LaViSE. As shown in the examples, we observe that LaViSE can explain convolutional filters by finding more accurate concepts that have not been trained. This suggests that our LaViSE can be deployed to any unannotated datasets to gain insights based on the explanations.

\begin{figure}[t]
    \centering
    \vspace{-10pt}
    \includegraphics[width=0.7\columnwidth]{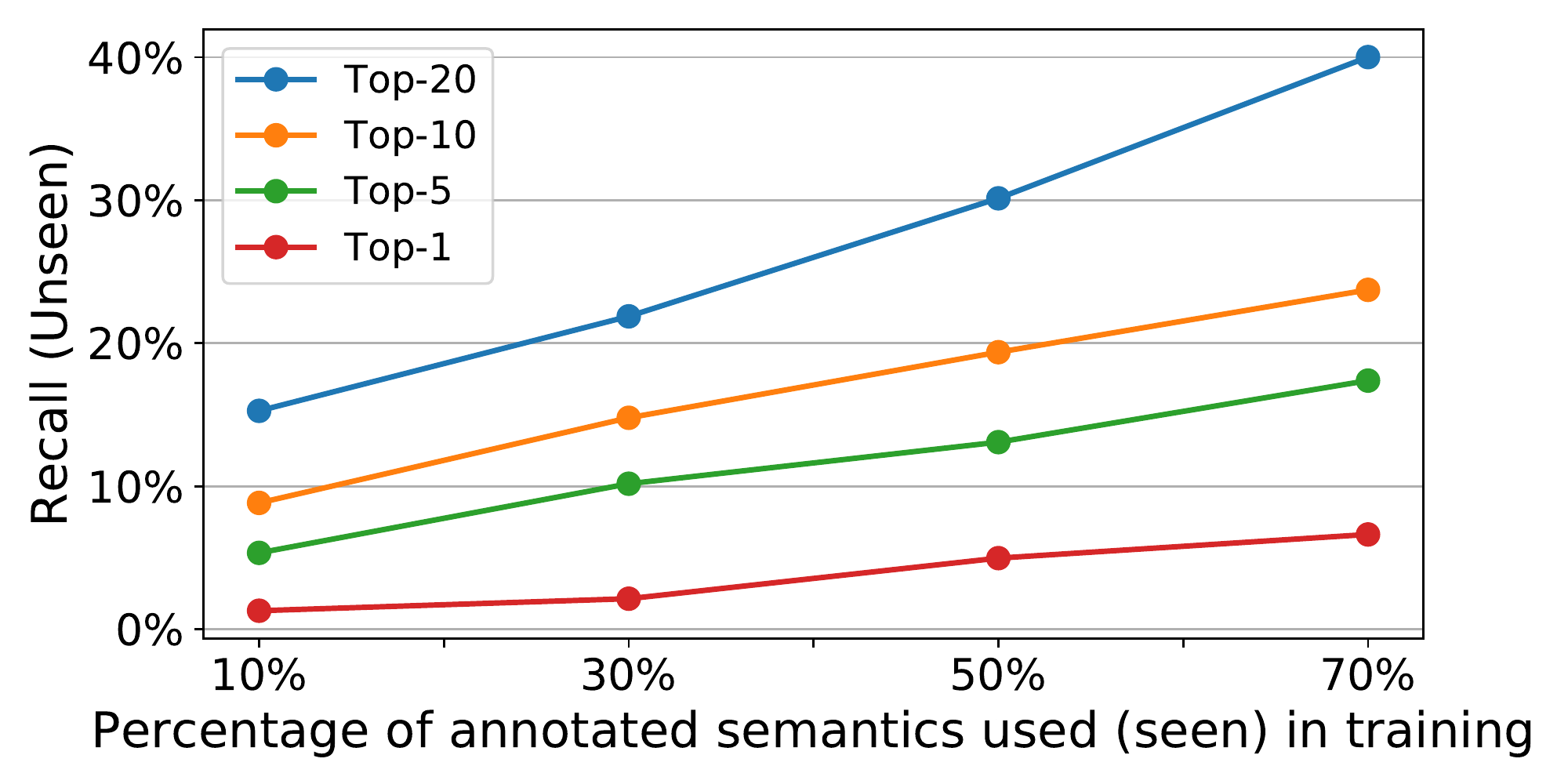}
    \vspace{-5pt}
    \caption{The proportion of the novel (unseen) concepts discovered by LaViSE.}
    \label{fig:novel}
    \vspace{-8pt}
\end{figure}

Figure~\ref{fig:novel} shows the proportion of the novel (i.e., unseen) concepts discovered by LaViSE according to the different percentages of annotated sementics in training. Note that we consider a concept as a novel concept when it matched to the annotated categories but not included in the training. We find that the recall values proportionally increase to the annotation rates in training. That is, LaViSE discovers more unseen concepts with a comprehensive reference dataset.



\subsubsection{Effect of Pre-training on Interpretability}
To understand the effect of pre-training on the interpretability of LaViSE, we compare two models with the same architecture but different pre-training procedures. 
Note that existing methods including NetDissect are not suitable for this analysis since new concepts would not be captured from a pre-trained model while LaViSE naturally supports this because it can use any arbitrary network and dataset.
More specifically, we take two ResNet18 models trained with the MS COCO dataset for multi-class classifications where one model is pre-trained with ImageNet~\cite{deng2009imagenet} while the other model is randomly initialized. 
According to the filter explanations generated by the LaViSE, even though two models have comparable classification accuracies, the model finetuned after pretraining on the ImageNet learned more concepts (219) than the model trained with MS COCO from a random initialization (150). We present the full list of discovered concepts in the Supplementary Material.

\subsection{Unsupervised Dataset Bias Analysis}
\label{sec:gender}
We now demonstrate a novel application of our LaViSE framework for the purpose of \textbf{unsupervised} comparative data and model analysis. The main purpose of the analysis is to discover and explain differences between multiple datasets, or different subsets of a given dataset without any labels. Prior work has also shown biases between datasets using labels~\cite{torralba2011unbiased}, but we are interested in \textbf{interpreting} the differences in an unsupervised fashion. This approach allows us to examine hidden biases in datasets or media outlets using machine learning models~\cite{thomas2021predicting,xi2020understanding}. 

We consider gender representation bias in public datasets as our examples here. Recent studies have reported various gender biases in image datasets and computer vision models such as accuracy disparity~\cite{buolamwini2018gender,karkkainen2021fairface} or spurious correlation~\cite{zhao-EtAl:2017:EMNLP20173,chen2021understanding,joo2020gender}. We mainly consider the latter, i.e. how gender correlates with other unknown covariates in the dataset.\footnote{Some papers also consider causal or counterfactual model bias or explanations~\cite{goyal2019explaining,goyal2019counterfactual,joo2020gender}. Our main interest is to explain biases in a dataset. } For example, Zhao et al.~\cite{zhao-EtAl:2017:EMNLP20173} showed that in popular image datasets gender is associated with activities such as shopping for women and driving for men. This analysis is supervised and requires annotations on activities or object categories. In contrast, LaViSE can directly apply to a dataset without any additional labels (except gender) such that it can \textbf{discover} hidden biases on unknown factors. 

\begin{figure}[t]
\centering
\vspace{-10pt}
\includegraphics[width=0.85\columnwidth]{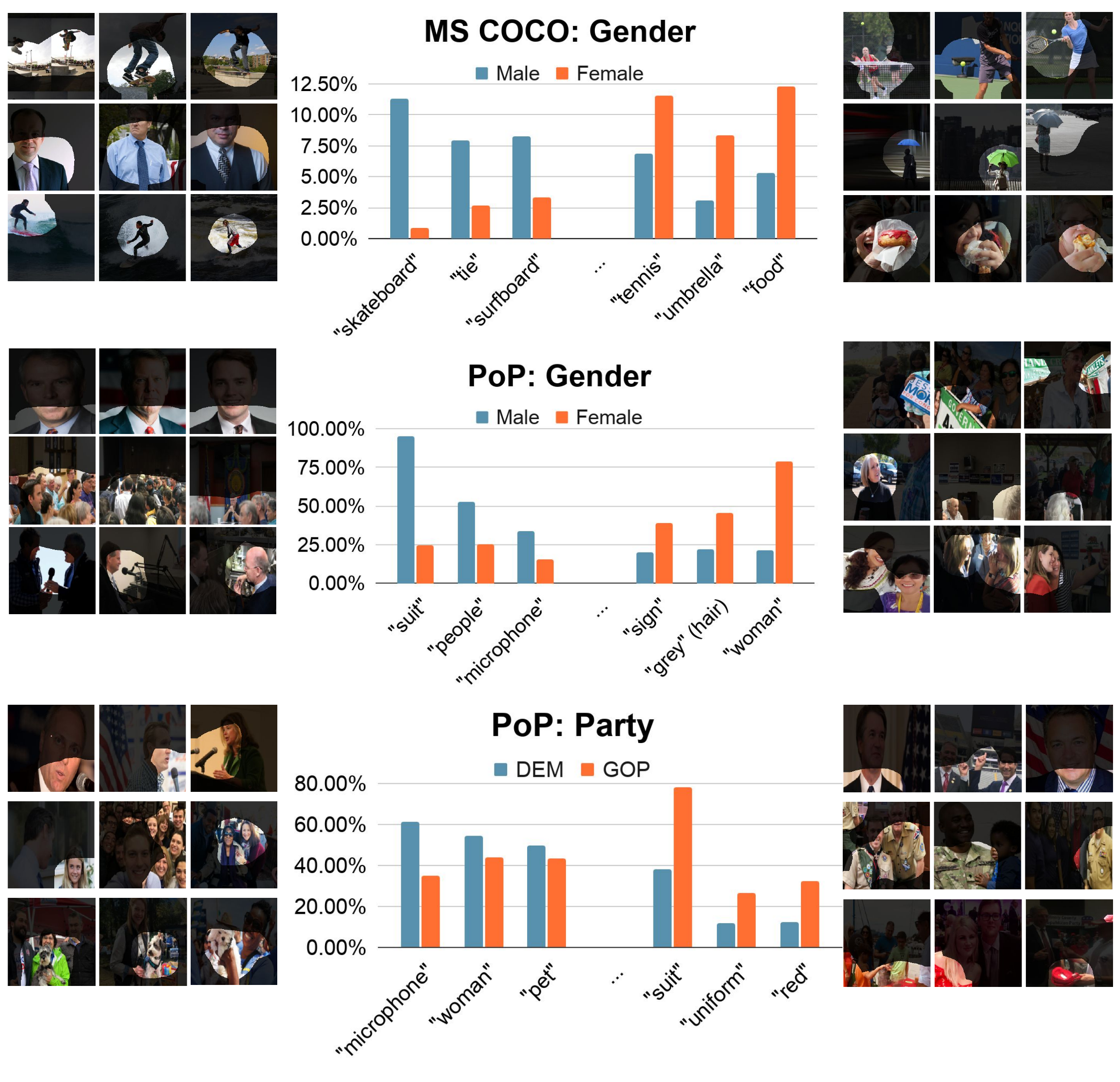}
\vspace{-10pt}
\caption{Comparative analysis of different groups of images in the MS COCO dataset and the PoP dataset.}
\label{fig:gender_coco}
\vspace{-15pt}
\end{figure}

Specifically, we show gender bias in the MS COCO dataset, i.e. which concepts are associated with gender. We split the images by gender according to \cite{zhao-EtAl:2017:EMNLP20173} and train a binary CNN (ResNet-18) classifying gender (we \textbf{only} use gender annotations). LaViSE then generates explanations for conv filters in the model. For each filter $u$, we count the number of images whose maximum activation is above the threshold $T_u$, and we call these images ``qualified images." Each gender is a group, and we compare the difference in percentages of qualified images between two groups. In Figure \ref{fig:gender_coco}, we list filters that have the most considerable distinctions between gender groups, their top-1 explanations predicted by LaViSE, and examples of qualified images on the sides. The results reflect the gender biases in this benchmark dataset and provide a guideline for future improvements of the dataset. Note that LaViSE can discover concepts which are \textbf{not} part of the COCO categories (e.g. food) as it does not use the categories or labels in analysis. 

We also use the same method to compare social media photographs of politicians between gender and party affiliations using PoP dataset. Such comparative analysis is essential in social science and media studies~\cite{bauer2018visual,lalancette2019power,wang2017polarized,thomas2021predicting,xi2020understanding,steinert2022state,unal2021visual,wang2016deciphering} but requires a huge amount of human effort for manual coding and may be susceptible to the bias of investigators. LaViSE offers an efficient data-driven way to explore group differences in unlabeled image datasets. The result in Figure \ref{fig:gender_coco} shows interesting gender and party differences. For example, male politicians tend to show large crowds to signal competence and popularity~\cite{joo2014visual,grabe2009image} and female politicians show more ``sign'' (panels) commonly used in public demonstrations and protests, which communicates their trustworthiness and interests in social welfare and protection for minority groups~\cite{everitt2016candidate,won2017protest}. 



\section{Conclusion}
We proposed LaViSE, a novel framework which can both visually and semantically explain latent representations of a trained CNN. It also enables users to discover concepts that a CNN learned without being explicitly taught. Empirical results show that our framework can accommodate different CNN architectures and datasets with varying formats of annotations. We also demonstrated a novel application for unsupervised bias analysis using our framework. We hope our work can help enhance transparency in both black-box models and datasets in AI research.

\noindent
\textbf{Acknowledgement.} This work was supported by NSF SBE-SMA \#1831848. 


\appendix
\newpage

\begin{strip}
\center
\textbf{\large 
    Explaining Deep Convolutional Neural Networks \\ via Latent Visual-Semantic Filter Attention: Appendix
}
\end{strip}
    

\begin{figure*}[t]
         \centering
         \includegraphics[width=0.7\textwidth]{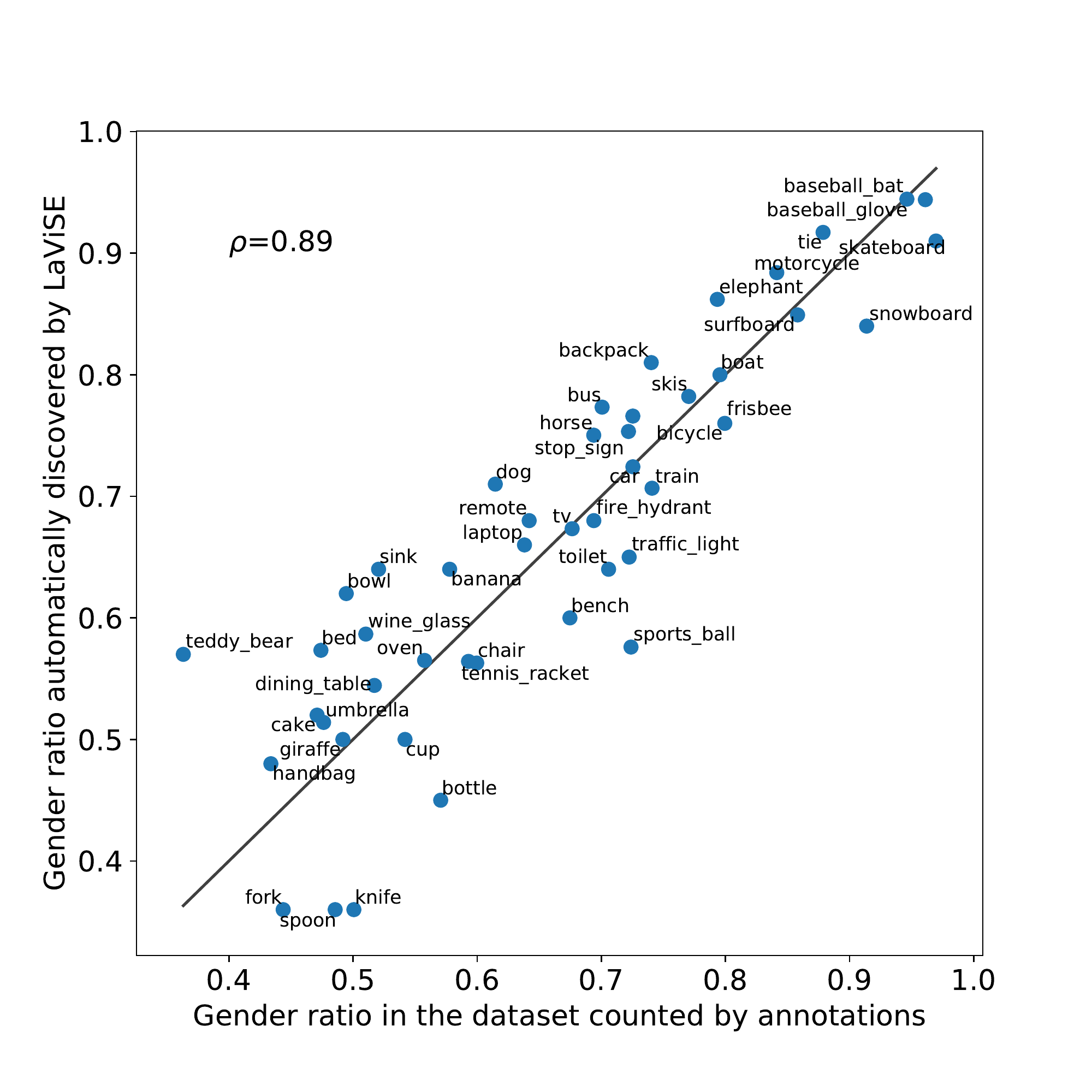}
     \caption{The validation of our unsupervised bias detection using MS-COCO annotations. The average gender biases discovered by our method are highly correlated with the actual gender ratios for the same concepts in the annotations of the dataset. $\rho$ is the Pearson correlation coefficient. 
    }
     \label{fig:gender_bias_coco}
\end{figure*}

\begin{figure*}[t]
\vspace{-10pt}
    \centering
    \includegraphics[width=0.62\textwidth]{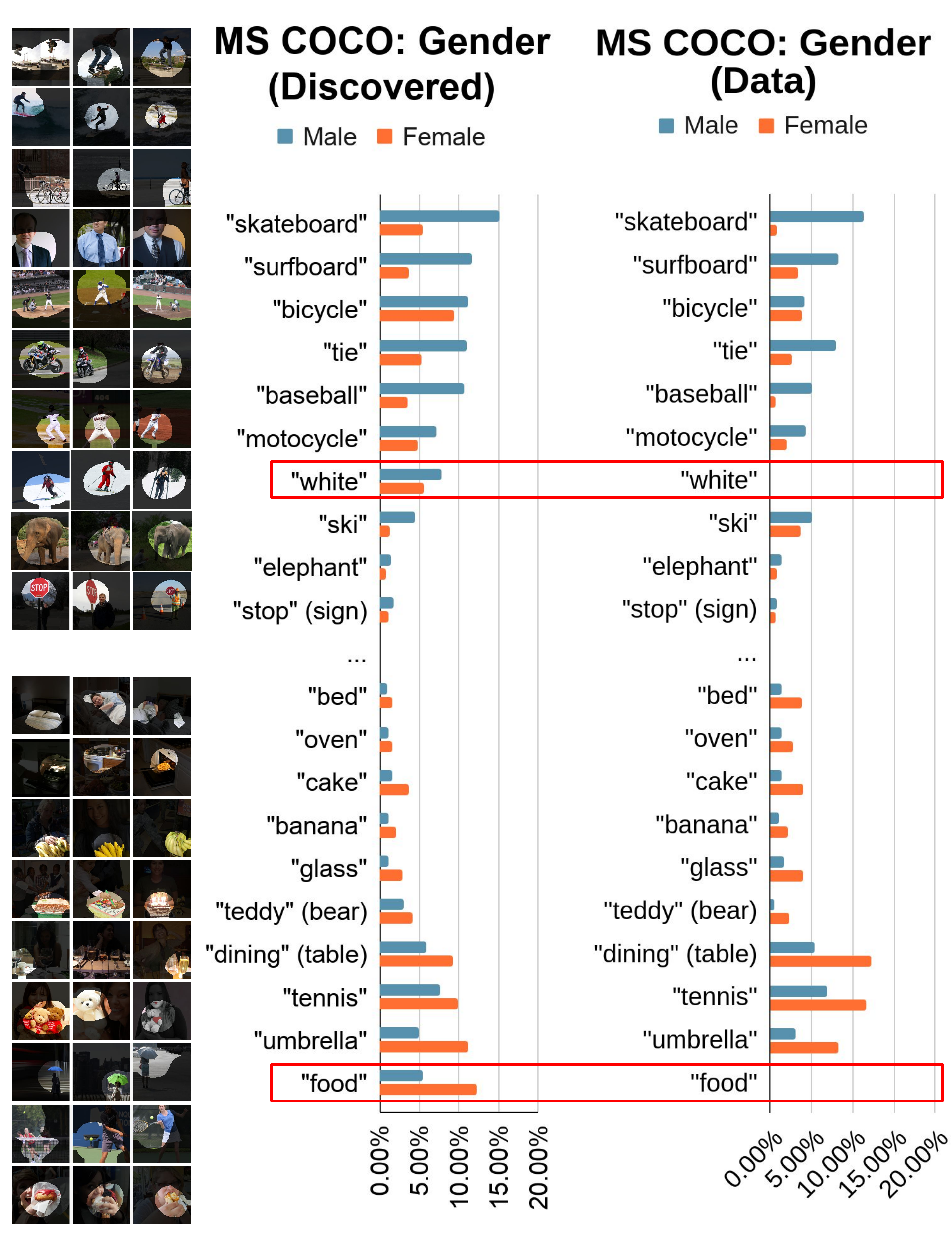}
    \caption{Concepts that best distinguish gender groups discovered by LaViSE from the last convolutional layer of a ResNet-18 model trained with the MS COCO dataset. ``White'' and ``food'' are not defined in the dataset but our method was able to find the concepts. For the ``white'' filter, our method also generates ``baseball'' and ``sleeve'' but ``white'' was the top word. This filter is also activated on the images of non-baseball players, which explains a smaller gender gap than the ``baseball'' filter. }
    \label{fig:coco_gender}
\end{figure*}

\begin{figure*}
    \begin{subfigure}[b]{0.35\textwidth}
    \centering
    \includegraphics[width=\columnwidth]{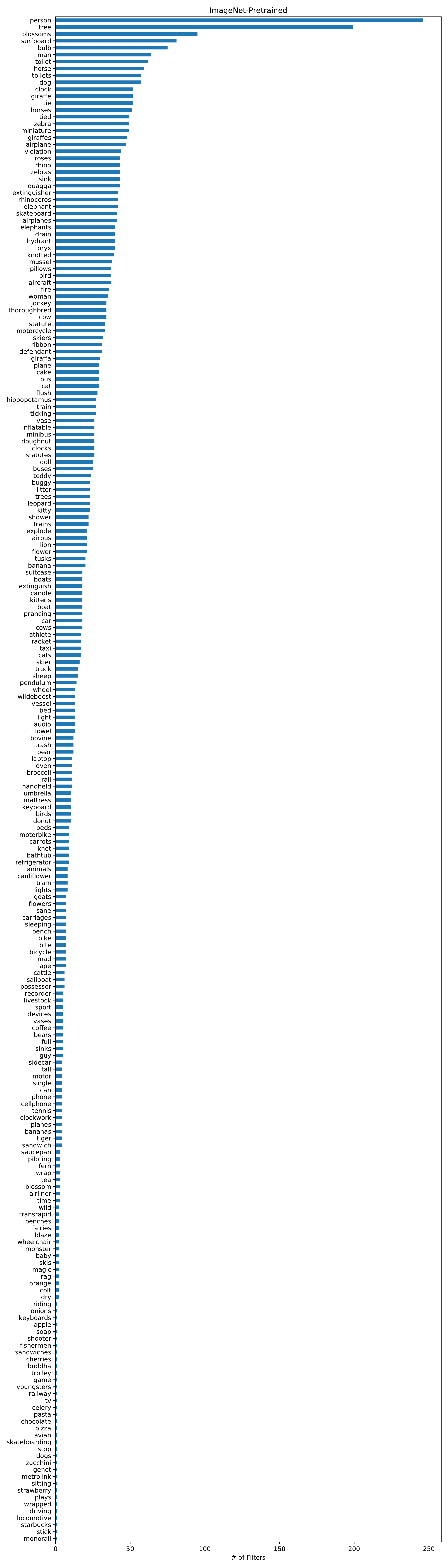}
    \caption{ImageNet-Pretrained}
    \label{fig:pretrain}
    \end{subfigure}
    \hfill
    \begin{subfigure}[b]{0.35\textwidth}
    \centering
    \includegraphics[width=\columnwidth]{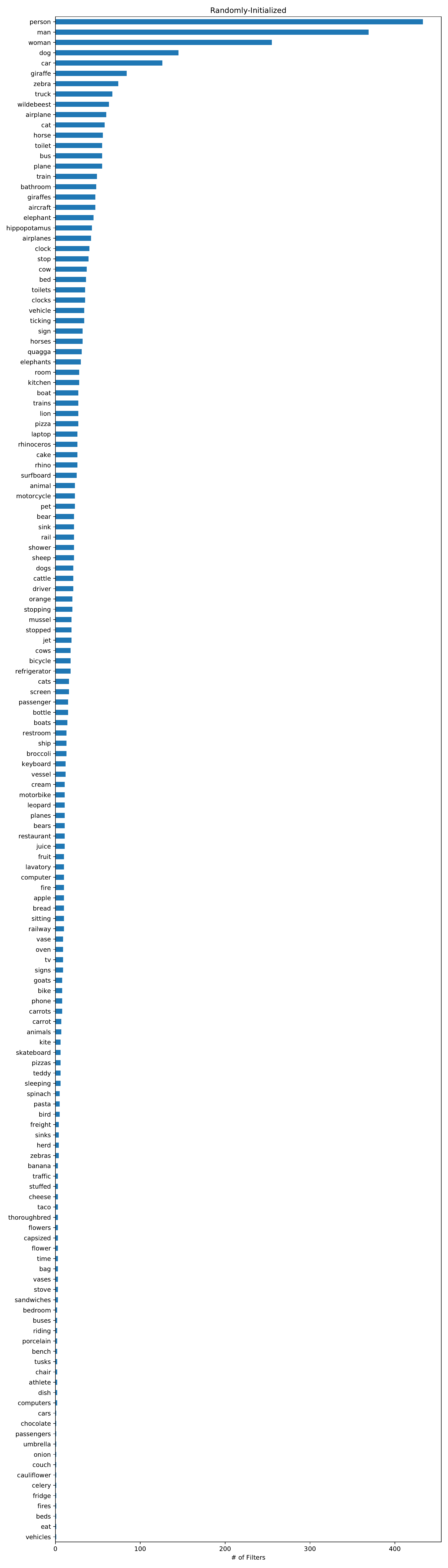}
    \caption{Randomly-Initialized}
    \label{fig:scratch}
    \end{subfigure}
    \caption{Comparative analysis of pretrained and randomly-initialized models trained with the MS COCO dataset.}
    \label{fig:concept_coco}
\end{figure*}

\section{Unsupervised Dataset Bias Analysis}

In Section~\ref{sec:gender}, we showed how LaViSE can be used for unsupervised bias analysis by discovering concepts that are associated with genders. Our method can uncover hidden biases without using any annotations except genders while prior studies~\cite{zhao-EtAl:2017:EMNLP20173} require human annotations. In this section, we present additional figures and examples to highlight the utility of our proposed framework.  



\paragraph{MS COCO is gender imbalanced.}
Our framework can be utilized to measure the gender imbalance of any dataset. In order to find out the gender balance in MS COCO~\cite{lin2014microsoft}, we use the same gender annotations for images in MS COCO as in \cite{zhao-EtAl:2017:EMNLP20173}. Following the metric presented in the same work, we compute the gender ratio of each object class $c$ in MS COCO as its bias toward males:
\begin{align}
    r_{gender, c} = \frac{N_{male,c}}{N_{male,c} + N_{female,c}}
\end{align}
where $N_{male,c}, N_{female,c}$ are the numbers of training images of class $c$ that also contain either males or females respectively, excluding samples that are related to both genders. 
Figure~\ref{fig:gender_bias_coco} shows the results of our unsupervised gender bias detection in the MS COCO dataset. Note that the x-axis indicates the gender ratios in the annotations of the dataset, and the y-axis shows the gender ratios discovered by LaViSE. A value above 0.5 indicates that there is a bias toward males for that class. We find that the gender ratios discovered by our method and in the dataset are highly correlated. This suggests that LaViSE can be applied to any dataset or trained model to detect biases without having additional annotations.

\paragraph{Our gender bias analysis is consistent with the gender ratios of the training data.} For each concept that we discover from the target model with LaViSE, we have at least one filter associated with this concept. To validate the explanations generated by LaViSE and the analysis supported by these explanations, we expect the gender ratios for the discovered concepts to have a positive correlation with the gender ratios of the corresponding classes in the dataset. For example, if we discover the ``tennis" concept from a trained model, we expect it to have a similar gender ratio as the ``tennis racket" class. Figure~\ref{fig:coco_gender} shows the concepts that best distinguish gender groups discovered by LaViSE in a ResNet-18 model trained on the MS COCO dataset. We find that LaViSE discovered some concepts that are not in the dataset. More specifically, as shown in Figure~\ref{fig:coco_gender}, the concepts ``white'' and ``food'' which have large gender gaps have been discovered by our method.

Note that the gender ratio for a concept at each associated filter $u$ is computed as the ratio of the number of male images to the sum of the number of male and female images. Formally,
\begin{align}
    r_{gender, u} = \frac{N_{male,u}}{N_{male,u} + N_{female,u}}
\end{align}
$N_{male,u}, N_{female,u}$ are the numbers of images for filter $u$ that include either males or females respectively. Then we simply take the average of the gender ratios of all associated filters to get the gender ratio for each concept.

\section{Details of the Evaluation Protocol}
To evaluate how accurate the explanation provided by the proposed framework is, we need to have ground-truth labels. However, there is no ground-truth labels for the concepts learned by each filter. In this section, we describe how LaViSE is evaluated without having the ground-truth labels.



Denote that $W_u$ is the composed explanations by using top-$\alpha$ ranked words for filter $u$, where $\alpha$ is a user-defined number.
For each filter $u$, it has $p$ most activated images $\{x_{i}^{u}\}_{i=1,...,p}$, and for each image $x_{i}^{u}$, we have its activated region 
\begin{align}
    R_{x_{i}^{u}} = (\text{Exp}(x_{i}^{u})_u > T_u)
\end{align}
on filter $u$. $T_u$ is a per-filter activation threshold and is determined in the same way as in \cite{zhou2018interpreting} such that $P(\text{Feat}(x_i) > T_u) = 0.005$ for all $x_i$.

Suppose $x_{i}^{u}$ is the only image that we will have to explain filter $u$. For each annotation mask $M_{j}$ of ${x_{i}^{u}}$, if the intersection-over-union score of $M_{j}$ and $R_{x_{i}^{u}}$ exceeds an universal threshold (0.04~\cite{zhou2018interpreting}), we consider the corresponding concept $t_{x_{i}^{u}, j}$ as a ground-truth concept for filter $u$. Then we denote $G_{u,i}$ as the ground-truth concepts for filter $u$.

Then for different choices of $\alpha$, our framework outputs a sequence of words $W_{u,q}$ ($|W_{u,q}|=\alpha$) to explain filter $u$. 
Based on the output words $W_{u,i}$ for explanation and the ground-truth concepts $G_{u,i}$, we finally compute the recall at filter $u$ as 
\begin{align}
    Recall_{u,i} = \frac{|W_{u,i} \bigcap G_{u,i}|}{|G_{u,i}|}
\end{align}
To compute the overall recall of using LaViSE to explain a target convolutional layer in a model, we take the average of the recalls given by different images and filters as follows:
\begin{align}
    Recall = \frac{1}{d\times p}\sum_{u=1}^d\sum_{i=1}^p Recall_{u,i}
\end{align}

\begin{table}[tbp]
\center
\footnotesize
\scalebox{0.9}{
\begin{tabular}{c|c|c|c}
\hline
Trained on & Recall (Top-5) & Recall (Top-10) & Recall (Top-20)\\
\hline
ImageNet & 0.226 & 0.302 & 0.373\\
\hline
Random init. & 0.009 & 0.025 & 0.034\\
\hline
\end{tabular}}
\caption{Interpretability on random-initialized feature extractor. 
\vspace{-10pt}
}
\label{tab:rand}
\end{table}

\section{Interpreting Random Features}
The objective of our method is to interpret any existing black-box models, not to learn more interpretable models or to interpret an uninterpretable model beyond what it actually captures. We design an experiment to test if our method can  only interpret what a model learned and does not generate irrelevant explanations. Foe example, a model that does not have any meaningful (interpretable) features may still yield interpretations by our method (\eg due to the explainer). To ensure that our method does not interpret uninterpretable features, we apply our method to a randomly initialized feature extrator.
Table~\ref{tab:rand} shows the result.
As expected, the interpretability is very low, validating that the explainer indeed explains only the features captured in the extractor. 





\section{Effect of the Number of Words}
\begin{figure}[!t]
     \centering
     \begin{subfigure}[b]{0.48\columnwidth}
         \centering
         \includegraphics[width=\columnwidth]{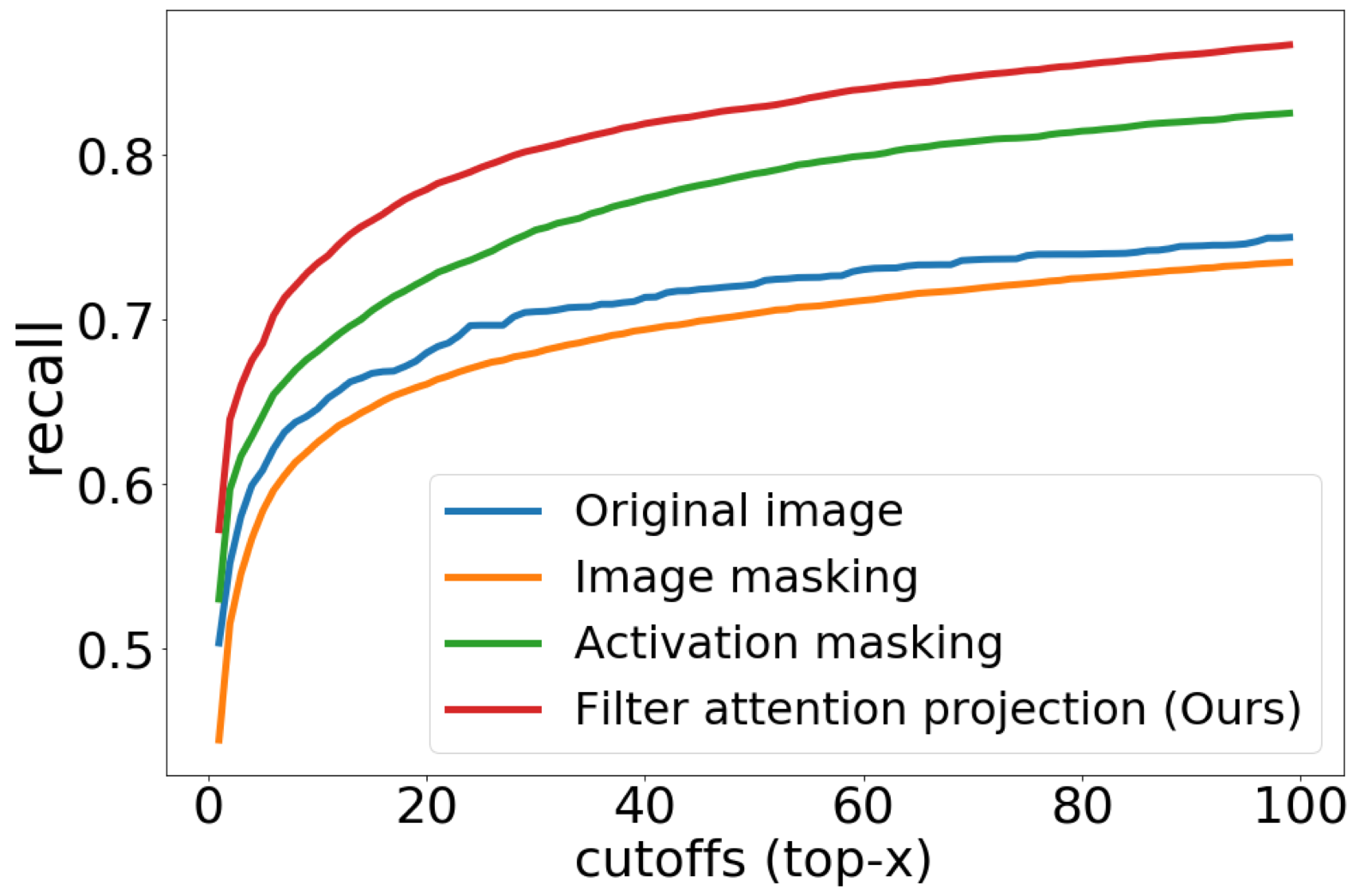}
         \caption{MSCOCO-ResNet18-Layer4}
     \end{subfigure}
     \hfill
     \begin{subfigure}[b]{0.48\columnwidth}
         \centering
         \includegraphics[width=\columnwidth]{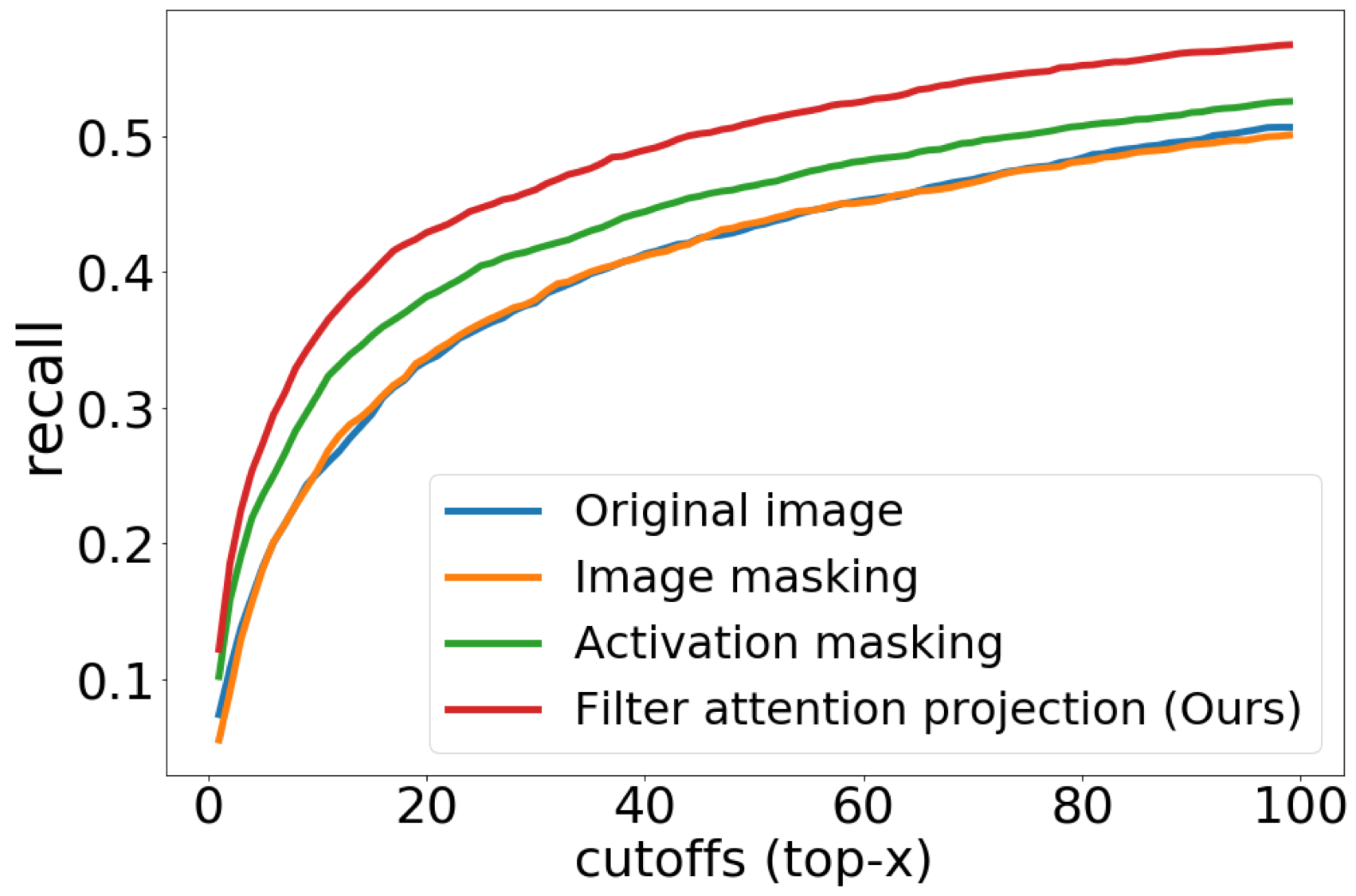}
         \caption{VG-ResNet18-Layer4}
     \end{subfigure}
     \begin{subfigure}[b]{0.48\columnwidth}
         \centering
         \includegraphics[width=\columnwidth]{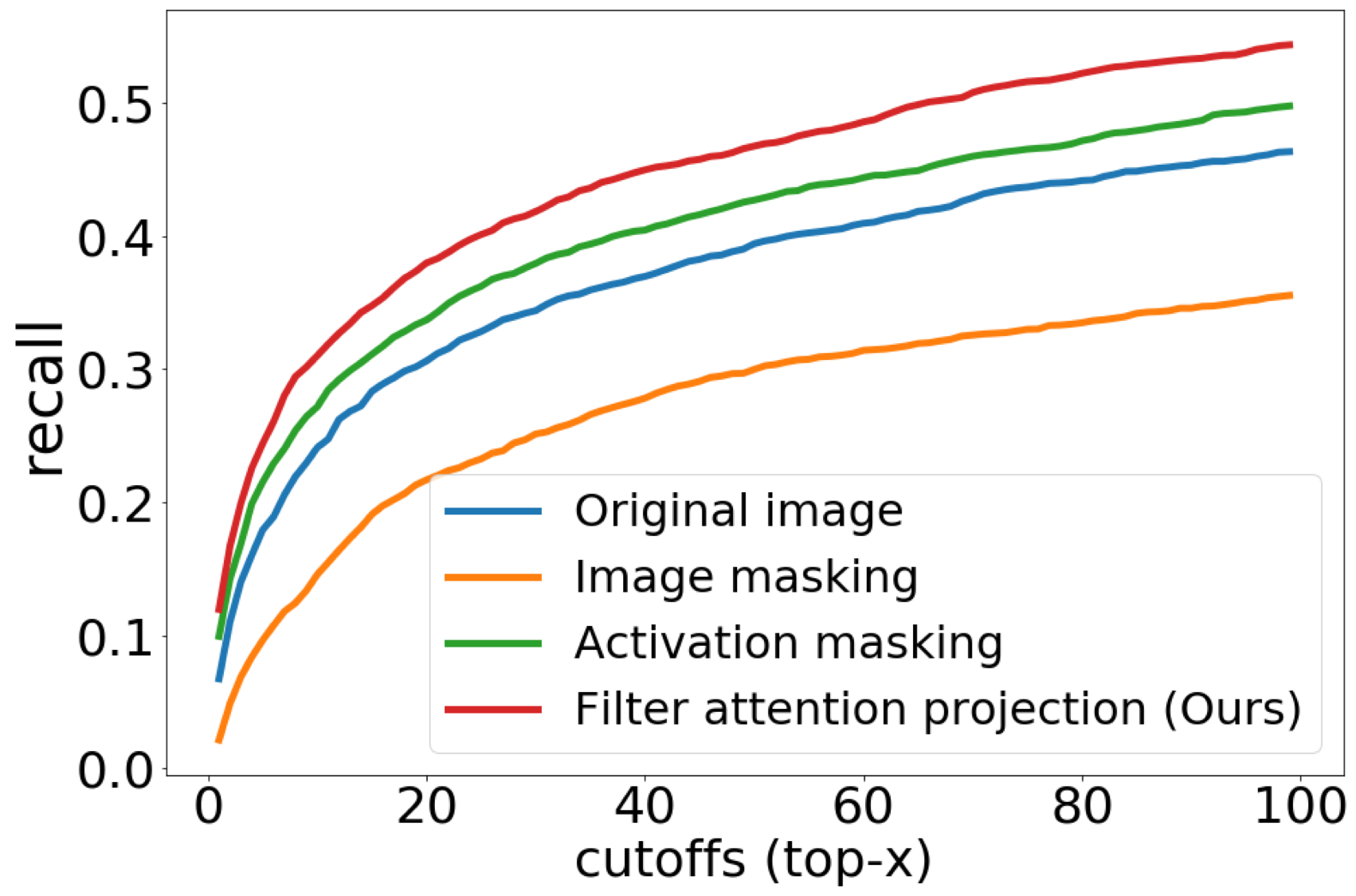}
         \caption{VG-ResNet50-Layer4}
     \end{subfigure}
     \hfill
     \begin{subfigure}[b]{0.48\columnwidth}
         \centering
         \includegraphics[width=\columnwidth]{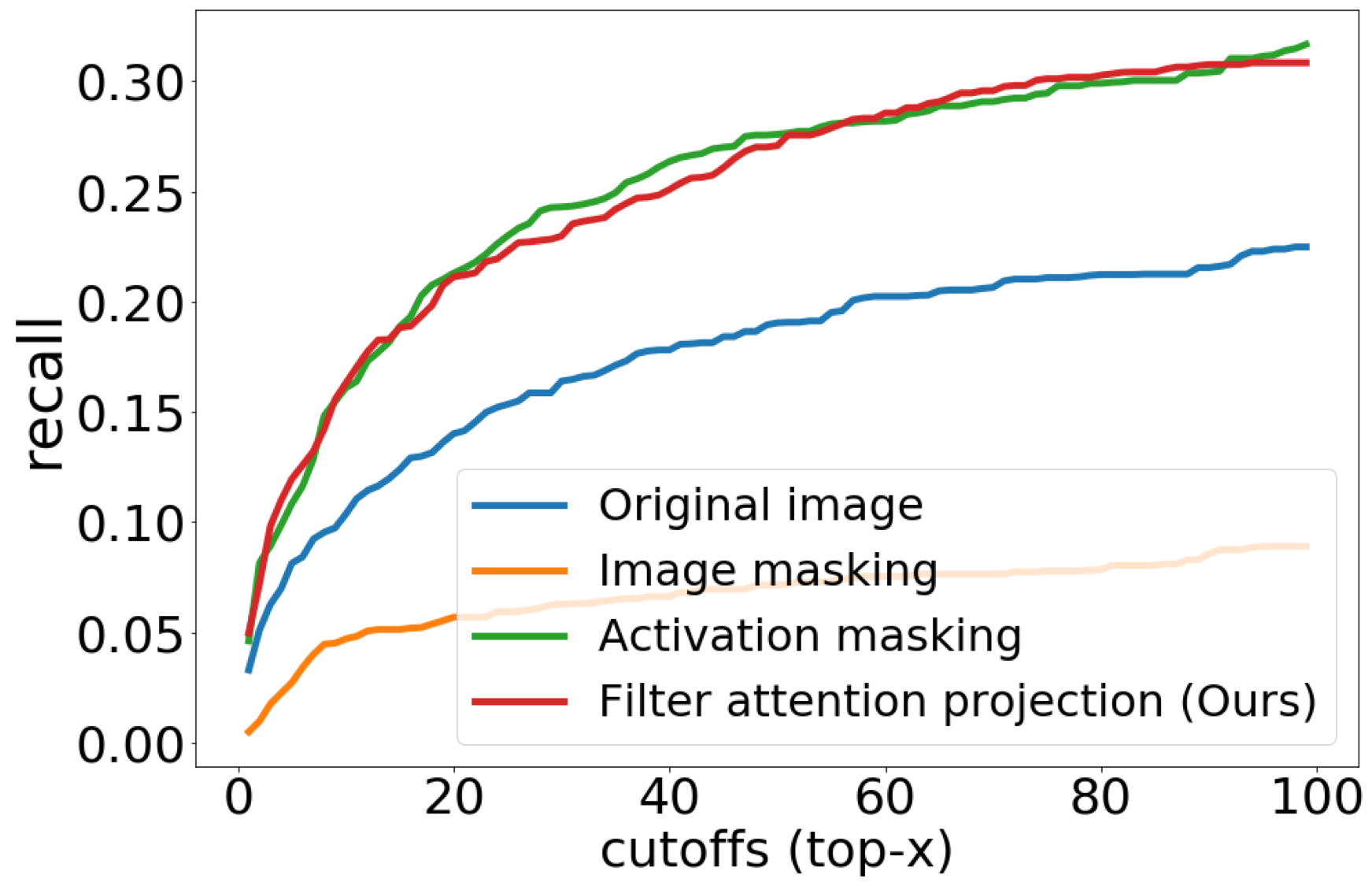}
         \caption{VG-ResNet50-Layer3}
     \end{subfigure}
\caption{Recall rates of three baseline methods and our framework when $x$ words are allowed for each explanation. These results correspond to the experiments shown in Table 2 in the main paper.}
\label{fig:coco_recall}
\end{figure}

Figure \ref{fig:coco_recall} shows the quantitative results based on alignment between explanations and annotations as an approximation of the baselines' performance when the problem scales (i.e., the number of words increases). Our method performs the best in all settings upon different numbers of words asked for the explanations. In the fourth case, although the activation masking baseline has a comparative performance with ours, it may not perform as well in reality as in this approximation while our method has stably good performance according to the results of human evaluation.

\section{Effect of Pre-training on Interpretability}
Following the discussion in Section 4.5.5, Figure \ref{fig:concept_coco} shows the full lists of concepts LaViSE used to explain a pretrained and a randomly-initialized model. Both models are ResNet-18 and trained with the MS COCO dataset. See Section 4.5.5 for details.

{\small
\bibliographystyle{ieee_fullname}
\bibliography{bib}
}

\end{document}